\documentclass[letterpaper, 10 pt, conference]{ieeeconf}  %

\IEEEoverridecommandlockouts                              %

\usepackage{footnpag}
\usepackage{graphicx}
\usepackage{amsmath}
\usepackage{amssymb}
\usepackage{bm}
\usepackage[version=4]{mhchem}
\usepackage{siunitx}
\usepackage{longtable,tabularx}
\usepackage{float}
\usepackage{caption}
\usepackage{booktabs} %
\usepackage{multirow}
\usepackage{adjustbox}
\usepackage[dvipsnames]{xcolor}
\usepackage{lipsum}
\usepackage{wrapfig}
\usepackage{diagbox}
\usepackage{algorithm}
\usepackage{cite}
\usepackage{hyperref}
\usepackage{cleveref}
\usepackage{doi}
\usepackage{censor}
\usepackage{tabularx}
\usepackage{array}
\hypersetup{
   colorlinks=true,
   allcolors=black,
   }

\makeatletter
\long\def\@makecaption#1#2{%
  \small  %
  \setbox\@tempboxa\hbox{#1.~#2}%
  \ifdim \wd\@tempboxa >\hsize
    #1.~#2\par
  \else
    \hbox to\hsize{\hfil\box\@tempboxa\hfil}%
  \fi}
\makeatother

\newif\ifanonymize
\anonymizefalse %

\ifanonymize
  \usepackage{censor}
  \let\anon\censor
\else
  \newcommand{\anon}[1]{#1}
\fi

\title{\LARGE \bf
More than a Manipulator: Planning Propellant-Free \\ Attitude Maneuvers for Free-Floating Spacecraft
}

\author{
\anon{Harsh G. Bhundiya$^{1}$, Avi Soval$^{2}$, and Keenan Albee$^{3}$}
\thanks{\anon{*This work was supported by the University of Southern California.}}
\thanks{\anon{$^{1}$Department of Astronautical Engineering,} \anon{University of Southern California,} \anon{Los Angeles, CA 90089;}
\anon{currently Assistant Professor, Department of Aerospace Engineering,} 
\anon{University of Maryland, College Park, MD 20742} \anon{{\tt\small bhundiya@umd.edu}}}
\thanks{\anon{$^{2}$Department of Aerospace and Mechanical Engineering,} \anon{University of Southern California,} \anon{Los Angeles, CA 90089} \anon{{\tt\small soval@usc.edu}}} 
\thanks{\anon{$^{3}$Departments of Astronautical, Aerospace and}
\anon{Mechanical, and Electrical and Computer Engineering,} \anon{University of Southern California,} \anon{Los Angeles, CA 90089} \anon{{\tt\small kalbee@usc.edu}}}
}

\begin{document}

\maketitle
\thispagestyle{empty}
\pagestyle{empty}

\begin{abstract}
Spacecraft attitude control is traditionally achieved using momentum exchange devices or propellant-consuming thrusters. Meanwhile, a growing number of missions require robotic manipulators, which are typically treated as disturbance sources to be rejected rather than as actuators for spacecraft reorientation. This work investigates the use of manipulator motions for propellant-free attitude control by formulating a trajectory optimization problem with critical joint and collision avoidance constraints. Using an interior-point solver for the resulting nonlinear program, complex slew and detumble trajectories are computed for a range of spacecraft-manipulator systems, demonstrating rest-to-rest slew maneuvers of up to 90$^\circ$ and detumble maneuvers with a 40\% reduction in angular velocity. The achievable control authority is also compared directly with that of reaction wheel arrays via momentum and torque envelopes, demonstrating the potential for manipulators to serve as redundant or even primary attitude control systems. This work provides a framework for using manipulators as multipurpose attitude control actuators, with particularly promising applications in in-space assembly and manufacturing when grasping payloads with high relative mass fractions.

\end{abstract}

\section{Introduction}\label{sec:intro}
Attitude control is essential for spacecraft operations but frequently relies on propellant-consuming thrusters for reorientation and momentum desaturation. Propellant-free attitude control is an enabler of long-duration missions and can provide a backup means of control when conventional actuators degrade or fail. Common methods include momentum exchange devices such as reaction wheels and control moment gyroscopes~\cite{leveSpacecraftMomentumControl2015}, and the exploitation of passive environmental disturbances including magnetic fields, atmospheric drag, and solar radiation pressure~\cite{atchison2011length}.

Internal mass reconfiguration, or the movement of masses relative to the spacecraft, presents an alternative attitude control approach with the potential for large control torques and low power consumption. Prior works have investigated large-angle slew maneuvers and detumbling through translation of internal masses~\cite{kane1969method,edwards1974automatic}, as well as combining internal mass motion with other actuators and environmental disturbances for attitude control of solar sails and variable-geometry spacecraft~\cite{wie2004solar,watanabe2023attitude}.

\begin{figure}[H]
    \centering
    \includegraphics[width=0.36\textwidth]{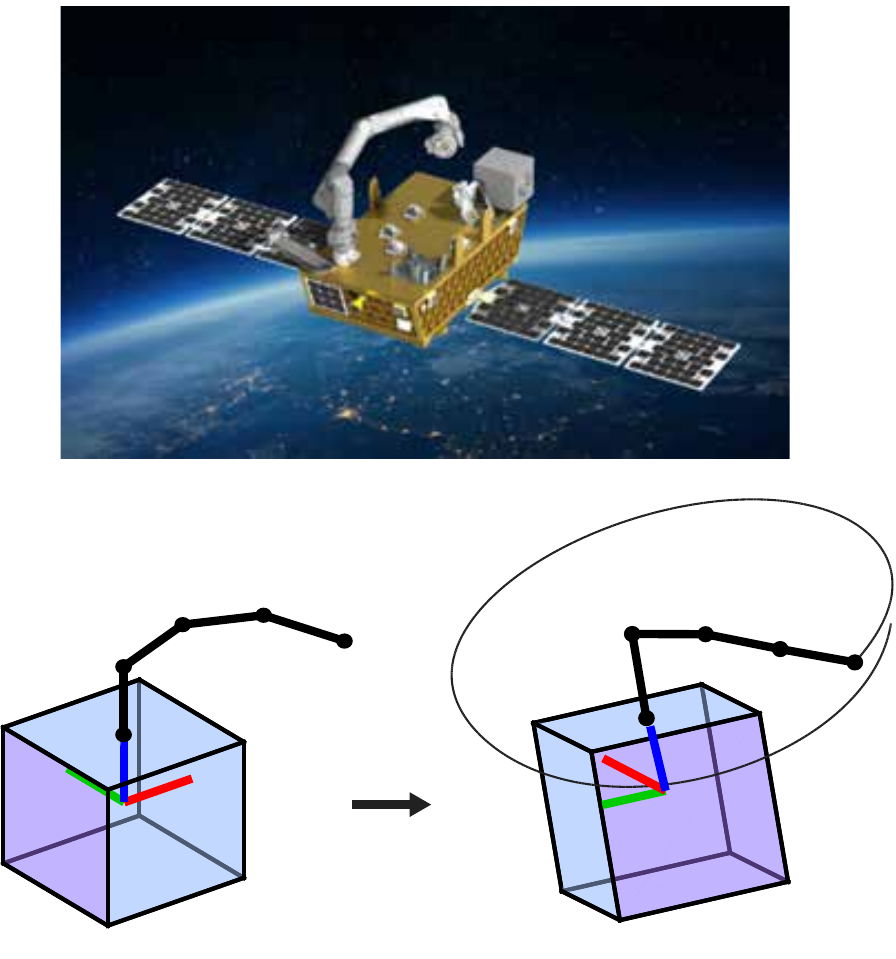}
    \caption{Robotic manipulator on NASA's Fly Foundational Robots mission~\cite{NASAFFR} (\textit{top}); a propellant-free slew maneuver of $75^\circ$ using solely manipulator motions (\textit{bottom}).}
    \label{fig:HookFigure}
\end{figure}

Robotic space manipulators, increasingly required for servicing, assembly, and debris-removal missions (Fig.~\ref{fig:HookFigure}), constitute another form of internal mass reconfiguration. Manipulator motions induce reactions of the base spacecraft through conservation of angular momentum; consequently, much of the existing literature has focused on minimizing the induced attitude disturbances during manipulation tasks. Early techniques understood how to minimize base attitude disturbances directly from the equations of motion via disturbance maps~\cite{torres1992minimizing}, and recent techniques have used offline trajectory optimization to compute joint trajectories that minimize the final base reorientation~\cite{crain2019experimental}.

Rather than treating manipulator-induced attitude motions as disturbances to be rejected, the same dynamic coupling can be exploited for spacecraft attitude control. This capability would allow robotic manipulators to serve as multipurpose systems, potentially reducing reliance on dedicated attitude actuators and enabling continued maneuvering after actuator saturation or failure. However, generating manipulator motions that achieve desired attitude maneuvers remains challenging due to the nonlinear coupled dynamics, high-dimensional search space, and stringent collision constraints.

In this paper, we focus on planning collision-free manipulator motions for attitude maneuvers of free-floating spacecraft-manipulator systems. Our specific contributions include:
\begin{itemize}
    \item A versatile optimal control problem (OCP) formulation with joint and collision avoidance constraints for attitude maneuvers using manipulator motions.
    \item Application of this OCP to slew and detumble maneuvers for a set of representative spacecraft-manipulator systems with up to 7 degrees of freedom.
    \item Comparison of manipulator motions to conventional momentum exchange devices in terms of attitude control authority and power consumption metrics.
\end{itemize}

The paper is organized as follows: Sec.~\ref{sec:relatedwork} surveys prior related work, Sec. \ref{sec:problemformulation} formulates the OCP, and Sec. \ref{sec:attitudemaneuvers} demonstrates propellant-free attitude maneuvers for example configurations. Sec. \ref{sec:RWcomparison} then compares the attitude control authority from manipulator motions to traditional momentum exchange devices, and Sec.~\ref{sec:conclusion} concludes the paper.

\section{Related Work} \label{sec:relatedwork}

\subsection{In-Flight Attitude Control via Internal Motions}

Many animals rely on internal motions for in-flight attitude control. Classic examples include falling cats and geckos which reorient during free fall to land on their feet through coordinated body and tail motions~\cite{kane1969dynamical}. Astronauts also routinely reorient in microgravity through body motions~\cite{passerelloHumanAttitudeControl1971}. Inspired by these examples, many prior works have employed inertial appendages to achieve in-flight attitude control on robotic systems. Early efforts focused on tailed robots such as Tailbot and RHex, where body reorientation was achieved through analytical models and feedback control laws that exploited angular momentum exchange between the tail and body~\cite{chang2011lizard,libby2016comparative}. Subsequent work has extended these ideas to reaction-wheel-equipped robots, where model-based controllers regulate wheel angular momentum to stabilize attitude during locomotion tasks~\cite{lee2023enhanced}. Chu et al.~\cite{chu2023combining} combined a tail and reaction wheel on a single platform and formulated the reorientation problem as a constrained quadratic program to achieve underactuated attitude control.

For systems with higher-dimensional appendages, direct model-based planning becomes increasingly challenging. Consequently, several works have employed learning-based approaches to discover coordinated inertial motions. Rudin et al.~\cite{rudin2022cat} trained reinforcement learning (RL) policies that use the legs of a quadruped as inertial appendages to perform cat-like aerial reorientation and landing in reduced-gravity environments. RL frameworks have also been investigated for combined locomotion, jumping, and in-flight attitude control in planetary exploration scenarios~\cite{olsen2026lowgravityplanetaryexplorationusing}. These works demonstrate that coordinated motions of internal appendages can provide substantial attitude authority even in highly nonlinear and underactuated systems, but without hard constraint enforcement.

The present work shares the same underlying principle of attitude control through internal mass motions, but considers free-floating spacecraft manipulators. Unlike terrestrial robotic applications that rely on learned policies or specialized feedback controllers, we formulate a constrained trajectory optimization problem that directly plans collision-free manipulator motions for spacecraft slew and detumble maneuvers, and explore comparisons against more traditional momentum exchange attitude control techniques.

\subsection{Dynamics and Motion Planning of Free-Floating Space Manipulators}
The dynamics of free-floating spacecraft with robotic manipulators have been extensively studied since the pioneering work of Dubowsky et al.~\cite{dubowskyKinematicsDynamicsControl1993}. A central characteristic of these systems is the dynamic coupling between manipulator motion and spacecraft attitude, whereby joint motions induce reactions on the base through conservation of linear and angular momentum. As a result, much of the early literature focused on understanding and mitigating manipulator-induced base disturbances during manipulation tasks. Representative examples include the virtual manipulator formulation of Vafa and Dubowsky~\cite{vafaDynamicsManipulatorsSpace1987} for small cyclic manipulator motions, the identification of dynamic singularities unique to free-floating systems by Papadopoulos and Dubowsky~\cite{papadopoulosDynamicSingularitiesFree1993}, and disturbance-minimization strategies based on null-space concepts and disturbance maps~\cite{torres1992minimizing}.

Motion planning for free-floating manipulators remains particularly challenging due to the nonlinear dynamics, high-dimensional search space, and need for collision avoidance. Nakamura et al.~\cite{nakamura1990nonholonomic} used a bidirectional approach to plan local base regulation trajectories about a reference point for a 6-DOF manipulator by exploiting the nonholonomic nature of the equations of motion. Tortopidis et al.~\cite{tortopidisPointtopointMotionPlanning2007} formulated the problem as a two-point boundary value problem and employed spline-based trajectory optimization for base disturbance-minimizing trajectories of planar free-floating systems. More recently, Crain et al.~\cite{crain2019experimental} applied pseudospectral optimal control methods to compute minimal disturbance trajectories for a planar spacecraft-manipulator system and experimentally validated the resulting motions. However, such approaches have primarily considered low-dimensional systems and end-effector motion objectives, with limited consideration of joint constraints, collision avoidance, and the ability to perform complex base attitude control maneuvers. 

Motivated by these limitations, this work uses the manipulator itself as an attitude control actuator and formulates a trajectory optimization problem that computes collision-free attitude maneuvers for general spacecraft-manipulator systems. In addition, the achievable attitude control authority is compared directly with conventional momentum exchange devices, providing insight into when manipulators may serve as redundant or primary attitude control actuators.

\section{Problem Formulation} \label{sec:problemformulation}

\subsection{System Dynamics} \label{sec:systemdynamics}
The rigid body dynamics of a free-floating spacecraft with an attached manipulator may be derived via the Newton-Euler or Euler-Lagrange formulations, as outlined in Refs.~\cite{crain2019experimental,dubowsky1991path}. Here, we briefly present the equations of motion to aid the optimal control problem formulation of Sec.~\ref{sec:OCP}.

\begin{figure}[t]
    \centering
    \includegraphics[width=0.275\textwidth]{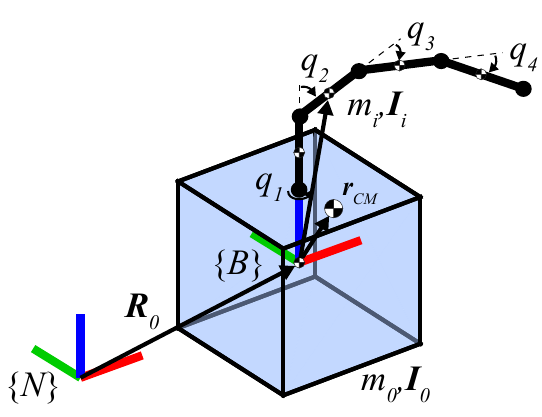}
    \caption{4-degree-of-freedom (DOF) manipulator arm on a free-floating spacecraft with defined parameters.}
    \label{fig:SCwFreeFloatingManipulator}
\end{figure}

As depicted in Fig.~\ref{fig:SCwFreeFloatingManipulator}, we consider a spacecraft-manipulator system of $n$ rigid bodies and $j$ revolute joints with angles $\bm{q}$ as specified using Denavit-Hartenberg (DH) parameters~\cite{lynch2017modern}. The linear and angular momenta of the combined system are~\cite{dubowsky1991path}
\begin{align}
\bm{P} &= \sum_{i=0}^{n} m_i\dot{\bm{R}}_i
       = M\dot{\bm{R}}_{CM},
\label{eq:P} \\
\bm{H} &= \tilde{\bm{I}}\bm{\omega}
       + \tilde{\bm{H}}.
\label{eq:H}
\end{align}

\noindent Here, $\tilde{\bm{I}}(\bm{q})$ represents the configuration-dependent inertia matrix of the combined system and $\tilde{\bm{H}}(\bm{q},\dot{\bm{q}})$ represents the instantaneous angular momentum from the manipulator motion relative to the base spacecraft:
\begin{align}
\tilde{\bm{I}} &= \sum_{i=0}^{n}\bm{I}_i - \sum_{i=1}^{n} m_i [\bm{r}_i-\bm{r}_{CM}]_\times [\bm{r}_i]_\times,
\label{eq:Itilda} \\
\tilde{\bm{H}} &= \sum_{i=1}^{n} \bm{I}_i\bm{\omega}_{i/B} + \sum_{i=1}^{n} m_i [\bm{r}_i-\bm{r}_{CM}]_\times \bm{r}'_i.
\label{eq:Htilda}
\end{align}

\noindent In Eqs. (\ref{eq:P})-(\ref{eq:Htilda}), $m_i$ is the mass of body $i$; $M=\sum_{i=0}^{n}m_i$ is the total system mass; $\bm{R}_i$ and $\bm{R}_{CM}$ are the inertial positions of body $i$ and the system center of mass, respectively; $\bm{r}_i$ and $\bm{r}_{CM}=\sum_{i=0}^{n}m_i\bm{r}_i/M$ are the position vectors from the body frame origin to the center of mass of body $i$ and the system center of mass, respectively; $\bm{I}_i$ is the inertia tensor of body $i$ about its center of mass; $\bm{\omega}$ is the angular velocity of the body frame {$\{B\}$} relative to the inertial frame {$\{N\}$}; $\bm{\omega}_{i/B}$ is the angular velocity of link $i$ relative to the body frame; $\bm{r}'_i$ is the velocity of the center of mass of body $i$ relative to the body frame; and $[\cdot]_\times$ denotes the skew-symmetric matrix corresponding to the cross product.

In this paper, we consider a free-floating environment with zero external forces and moments, resulting in conservation of linear and angular momentum. The equations of motion are thus obtained by differentiating Eqs. (\ref{eq:P})-(\ref{eq:H}) via the transport theorem and equating to zero, i.e., 
\begin{align}
\dot{\bm{P}} &= M\ddot{\bm{R}}_{CM} = 0,
\label{eq:Pconserve} \\
\dot{\bm{H}} &= \frac{d\bm{H}}{dt} + [\bm{\omega}]_\times \bm{H} = 0.
\label{eq:Hconserve}
\end{align}

\noindent Equation~(\ref{eq:Pconserve}) implies that the system center of mass does not accelerate in the inertial frame, and Eq.~(\ref{eq:Hconserve}) governs the attitude dynamics of the combined spacecraft-manipulator system. It can be shown that Eq. (\ref{eq:Hconserve}) contains nonintegrable velocity constraints, creating nonholonomic dynamics that enable the base attitude to be modified by path-dependent joint trajectories \cite{nakamura1990nonholonomic}. 

\subsection{Optimal Control Problem}\label{sec:OCP}
To solve for the manipulator motions that result in the desired attitude maneuver of the base spacecraft, we formulate an optimal control problem (OCP) that seeks to reach a commanded state while satisfying constraints on the system dynamics, joint angle limits, and self-collision avoidance. 

% The OCP has $13+2N$ states: 
The OCP has $13+2j$ states: the system state $\bm{x}$ contains the position $\bm{R}_0\in\mathbb{R}^{3}$, orientation parameterized by the unit quaternion $\bm{q}_{BN}\in\mathbb{R}^4$, linear velocity $\dot{\bm{R}}_0\in\mathbb{R}^{3}$, angular velocity $\bm{\omega}\in\mathbb{R}^{3}$, as well as the joint angles and velocities $\bm{q},\dot{\bm{q}}\in\mathbb{R}^{j}$. The control inputs are the joint accelerations $\ddot{\bm{q}}\in\mathbb{R}^{j}$, i.e,
\begin{equation}
\begin{aligned}
    \bm{x} &=
    \begin{bmatrix}
    \bm{R_0} & 
    \bm{q}_{BN} & 
    \dot{\bm{R}_0} & 
    \bm{\omega} & 
    \bm{q} & 
    \dot{\bm{q}}
    \end{bmatrix}^\top
    \in \mathbb{R}^{13+2j}, \\
    \bm{u} &= \ddot{\bm{q}} \in \mathbb{R}^j.
\end{aligned}
\end{equation}

The objective is to determine joint acceleration profiles that drive the terminal state towards a desired configuration. To this end, the stage cost tracks the difference between the desired states $[\cdot]_d$ and current states $[\cdot]_k$ with the error 
\begin{equation}
    \bm{e}_k = 
    \begin{bmatrix}
        \bm{R}_{0,d}-\bm{R}_{0,k} \\
        \bm{e}_{vk} \\ 
        \dot{\bm{R}}_{0,d} - \dot{\bm{R}}_{0,k} \\
        \bm{\omega}_d - \bm{\omega}_k \\
        \bm{q}_d - \bm{q}_k \\ 
        \dot{\bm{q}}_d - \dot{\bm{q}}_k
    \end{bmatrix}
    \in \mathbb{R}^{12+2j},
\end{equation}
where $\bm{e}_{vk} = \operatorname{vec}\!\left(\bm{q}_{BN,d}^{-1} \otimes \bm{q}_{BN,k}\right)$ is the vector part of the error quaternion between $\bm{q}_{BN,d}$ and $\bm{q}_{BN,k}$. This three-parameter attitude error avoids the quaternion sign ambiguity, provides a smooth attitude error metric, and is locally equivalent to logarithmic-map and geodesic-angle error representations on $SO(3)$ for small attitude errors~\cite{sola2017quaternion}.

The discretized cost function over a horizon $N$ is then 
\begin{equation}\label{eq:cost}
    J = \sum_{k=0}^{N-1} \bm{u}_k^\top \bm{R}\, \bm{u}_k^{} + \bm{e}_N^\top \bm{Q} \bm{e}_N^{}
\end{equation}
\noindent where a quadratic stage cost penalizes control effort, with a terminal penalty on state error. The weight matrices are parametrized by $[w_u, w_r, w_{e_{vf}}, w_{\dot{r}},  w_\omega, w_q, w_{\dot{q}}]$:
\begin{gather}
\bm{R} = w_u \bm{I}_{j},
\label{eq:Rmatrix}\\
\bm{Q} = \operatorname{diag}
\bigl(
w_r\bm{I}_3,
w_{e_{vf}}\bm{I}_3,
w_{\dot r}\bm{I}_3,
w_\omega\bm{I}_3,
w_q\bm{I}_{j},
w_{\dot q}\bm{I}_{j}
\bigr).
\label{eq:Qmatrix}
\end{gather}

The key constraints on the state and control inputs relate to the joint limits and collision avoidance. In particular, for each joint $i$, angle and velocity limits are enforced over the horizon $k\in[0,N]$ and accelerations are bounded over $k\in[0,N-1]$:
\begin{gather}
q_{i,\min} \le q_{i,k} \le q_{i,\max},
\label{eq:jointAngleLimit}\\
|\dot{q}_{i,k}| \le \dot{q}_{i,\max},
\label{eq:jointRateLimit}\\
|u_{i,k}| \le u_{i,\max}.
\label{eq:jointAccelLimit}
\end{gather}

To prevent self-collisions and collisions between the manipulator and the base spacecraft, additional geometric constraints are incorporated. The rigid links that comprise the manipulator are modeled as zero-thickness line segments, and the distance function $\mathrm{dist}(\cdot)$ is computed using Euclidean distances, which can be inflated to a desired minimum clearance. To avoid spacecraft-manipulator collision, each link $\mathcal{L}_a$, except the first which is attached to the base, must maintain a minimum clearance $d_{\min}$ from a sphere of radius $R_{\mathrm{body}}$ circumscribing the base. Additionally, non-adjacent links must satisfy a minimum separation distance to prevent self-collisions:
\begin{gather}
\mathrm{dist}(\mathcal{L}_a,\mathcal{B})
\ge R_{\mathrm{body}}+d_{\min},
\quad a\ge2,
\label{eq:bodyCollision}\\
\mathrm{dist}(\mathcal{L}_a,\mathcal{L}_b)
\ge d_{\min},
\quad b\ge a+2.
\label{eq:selfCollision}
\end{gather}
Adjacent links are excluded from the self-collision check of Eq.~\ref{eq:selfCollision} when they share a joint and implicitly avoid collision from DH parameter selection.

Thus, the discrete OCP solved at each iteration is
\begin{alignat}{2}
    \min_{\{\bm{X},\bm{U}\}} \quad & J(\bm{X},\bm{U}) && \nonumber\\
    \text{s.t.} \quad
    & \bm{x}_0 = \bm{x}_0^{\mathrm{given}}, && \nonumber\\
    & \bm{x}_{k+1} = \bm{\Phi}_{\mathrm{RK4}}(\bm{x}_k,\bm{u}_k,\Delta t),
      & k \in [0,N{-}1], \nonumber\\
    & |u_{i,k}| \le u_{i,\max},
      & k \in [0,N{-}1], \nonumber\\
    & q_{i,\min} \le q_{i,k} \le q_{i,\max},
      & k \in [0,N], \nonumber\\
    & |\dot{q}_{i,k}| \le \dot{q}_{i,\max},
      & k \in [0,N], \nonumber\\
    & \mathrm{dist}(\mathcal{L}_a, \mathcal{B}) \ge R_{\mathrm{body}} + d_{\min},
      \; a \ge 2, & k \in [0,N], \nonumber\\
    & \mathrm{dist}(\mathcal{L}_a, \mathcal{L}_b) \ge d_{\min},
      \; b \ge a{+}2, & k \in [0,N]. \nonumber
\end{alignat}
where $\bm{\Phi}_{\mathrm{RK4}}$ denotes one step of the fourth-order Runge--Kutta integrator applied to the equations of motion of Sec.~\ref{sec:systemdynamics} and $\bm{X}=\{\bm{x}_k\}_{k=0}^{N} \in \mathbb{R}^{(13+2j) \times (N+1)}, \bm{U} = \{\bm{u}_k\}_{k=0}^{N-1} \in \mathbb{R}^{j \times N}$ constitute the set of decision variables.

This formulation is used for all subsequent trajectory computations. Direct multiple shooting transcribes the OCP into a sparse nonlinear program (NLP) whose decision variables are the base and joint states at each shooting node, and the joint accelerations over each control interval. Joint, actuation, and collision constraints are imposed at each node, and the continuous-time dynamics are enforced as dynamic-consistency equality constraints between successive nodes using fixed-step RK4 integration. This NLP is constructed in CasADi~\cite{Andersson2019}, which supplies first- and second-order derivatives of the dynamics, constraints, and objective function via automatic differentiation. The resulting NLP is solved with IPOPT, a primal-dual interior-point method that takes iterative steps to satisfy KKT optimality conditions~\cite{wachter2006implementation}, yielding a locally optimal solution trajectory consistent with the discretized dynamics and imposed constraints. 

\section{Propellant-Free \\ Slew and Detumble Maneuvers}\label{sec:attitudemaneuvers}

Using the OCP of Sec.~\ref{sec:problemformulation}, we now demonstrate propellant-free attitude maneuvers of the base spacecraft using manipulator motions. Two fundamental maneuvers are considered: (i) a slew with zero angular velocity at the initial and final states, and (ii) a detumble that reduces an initially nonzero angular velocity. Slew maneuvers are routinely required in spacecraft operations to reorient payloads or align antennas for communication, and detumble maneuvers are often required following deployment, actuator failures, or emergency scenarios that result in undesired rotational motion.

To compute manipulator motions for these maneuvers, we encode the desired final states in the OCP and accordingly adjust the terminal cost weights. For slew maneuvers, the goal is to achieve a desired final orientation $\bm{q}_{BN,d}$ of the base with no residual angular or joint velocity. In contrast, detumble maneuvers specify only an initial angular velocity $\bm{\omega}_0$, and the OCP terminal cost seeks to minimize the final angular velocity $\bm{\omega}_f$ and joint velocities $\dot{\bm{q}}$. Neither case specifies final joint angles, ensuring the manipulator can move freely to satisfy the desired maneuver.

\begin{figure}[t]
    \centering
    \includegraphics[width=0.45\textwidth]{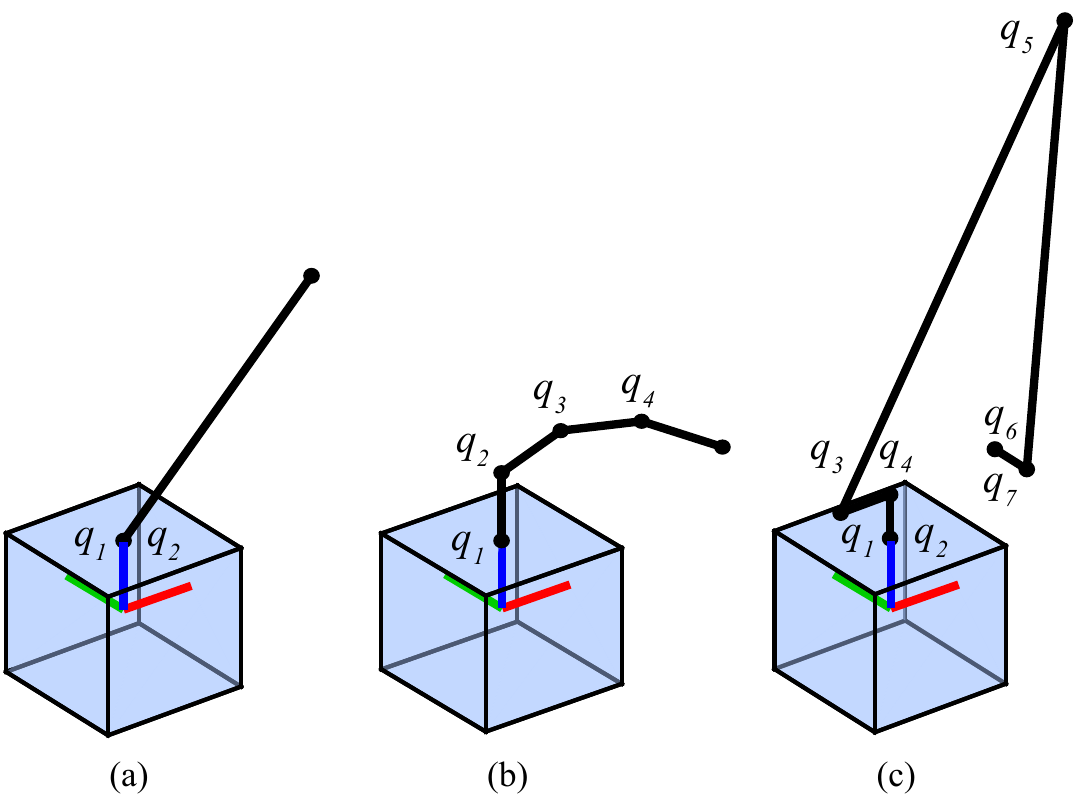}
    \caption{Spacecraft-manipulator systems with increasing degrees of freedom: (a) 2-DOF manipulator, (b) 4-DOF manipulator representative of the xLink robotic arm system~\cite{xLink}, 7-DOF manipulator representative of the Candarm2~\cite{nokleby2007singularity}.}
    \label{fig:ManipulatorConfigs}
\end{figure}

\begin{table*}[htp]
\centering
\caption{Mass properties, DH parameters, and joint constraints for the spacecraft-manipulator systems of Fig.~\ref{fig:ManipulatorConfigs}.}
\vspace{7pt}
\label{tab:systemparams}
\footnotesize
\setlength{\tabcolsep}{4pt}
\begin{tabular}{c c c c c c c c c c c c c c}
&
\multicolumn{2}{c}{Spacecraft} &
\multicolumn{3}{c}{Links} &
\multicolumn{4}{c}{DH Parameters} &
\multicolumn{4}{c}{Joint Constraints} \\
\cmidrule(lr){2-3}\cmidrule(lr){4-6}\cmidrule(lr){7-10}\cmidrule(lr){11-14}
& $m_0$ & $s$ & $i$ &
$m_i$ & $\sum m_i$ &
$a_i$ & $\alpha_i$ & $d_{0i}$ & $\theta_{0i}$ &
$q_{i,\min}$ & $q_{i,\max}$ & $\dot{q}_{i,\max}$ & $u_{i,\max}$ \\
& [kg] & [m] & &
[kg] & [kg] &
[m] & [deg] & [m] & [deg] &
[deg] & [deg] & [deg/s] & [deg/s$^2$] \\
\hline
\multirow{2}{*}{(a)} & \multirow{2}{*}{$300$} & \multirow{2}{*}{$0.5$}
  & 1 & $0$  & \multirow{2}{*}{$25$}   & $0$   & $-90$ & $0$ & $0$      & $-\infty$ & $+\infty$
    & \multirow{2}{*}{$17.2$} & \multirow{2}{*}{$11.5$} \\
  &&& 2 & $25$ & & $1.0$ & $0$      & $0$ & $-90$ & $-90$     & $+90$ & & \\
\hline
\multirow{4}{*}{(b)} & \multirow{4}{*}{$300$} & \multirow{4}{*}{$0.5$}
  & 1 & $7$ & \multirow{4}{*}{$25$}    & $0$    & $90$ & $0.25$ & $-180$  & $-\infty$ & $+\infty$
    & \multirow{4}{*}{$17.2$} & \multirow{4}{*}{$11.5$} \\
  &&& 2 & $6$ & & $0.25$ & $0$    & $0$    & $90$ & $-90$     & $+90$  & & \\
  &&& 3 & $6$ & & $0.25$ & $0$    & $0$    & $0$     & $-50$     & $+50$  & & \\
  &&& 4 & $6$ & & $0.25$ & $0$    & $0$    & $0$     & $-35$     & $+35$  & & \\
\hline
\multirow{7}{*}{(c)} & \multirow{7}{*}{$400{,}000$} & \multirow{7}{*}{$2.0$}
  & 1 & $0$   & \multirow{7}{*}{$1{,}500$} & $0$     & $0$      & $0$     & $0$ & $-180$ & $+180$
    & \multirow{7}{*}{$17.2$} & \multirow{7}{*}{$11.5$} \\
  &&& 2 & $60$  & & $0$     & $90$  & $0.635$ & $0$ & $-180$ & $+180$ & & \\
  &&& 3 & $70$  & & $0$     & $-90$ & $0.756$ & $0$ & $45$     & $+150$    & & \\
  &&& 4 & $620$ & & $6.85$ & $0$      & $0$     & $0$ & $-45$     & $+45$    & & \\
  &&& 5 & $620$ & & $6.85$ & $0$      & $0.756$ & $0$ & $30$    & $+150$    & & \\
  &&& 6 & $70$  & & $0$     & $90$  & $0.635$ & $0$ & $-180$    & $+180$    & & \\
  &&& 7 & $60$  & & $0$     & $-90$ & $0$     & $0$ & $-180$    & $+180$    & & \\
\hline
\end{tabular}
\end{table*}

\begin{figure*}[htp]
    \centering
    \includegraphics[width=1.0\textwidth]{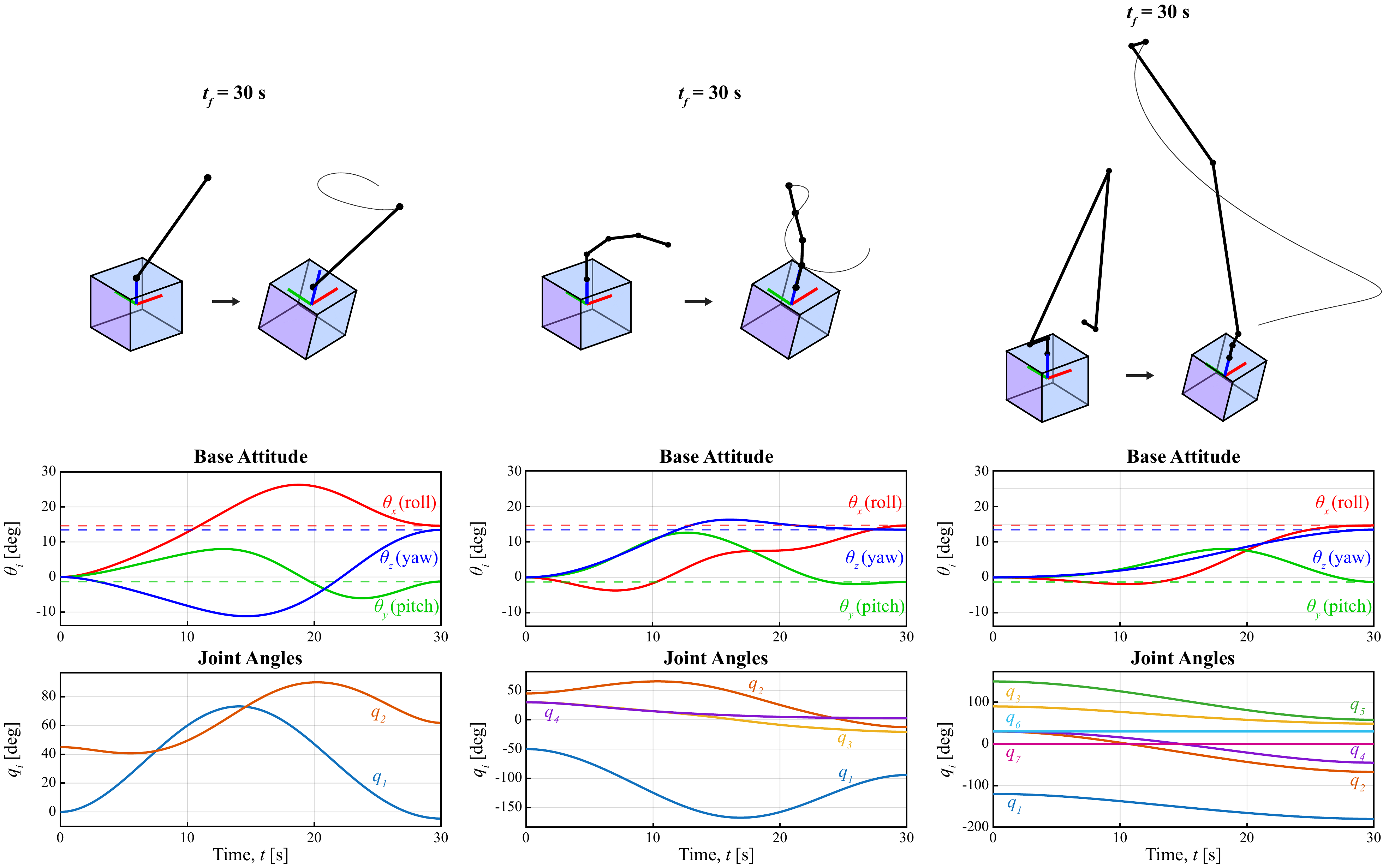}
    \caption{
        Propellant-free slew maneuvers of $\theta = 20^\circ$ for the spacecraft-manipulator systems of Fig.~\ref{fig:ManipulatorConfigs}. Manipulator motions avoid collision and satisfy the imposed joint constraints.}
    \label{fig:MegaSlewFig}
\end{figure*}

\begin{figure*}[htp]
    \centering
    \includegraphics[width=1.0\textwidth]{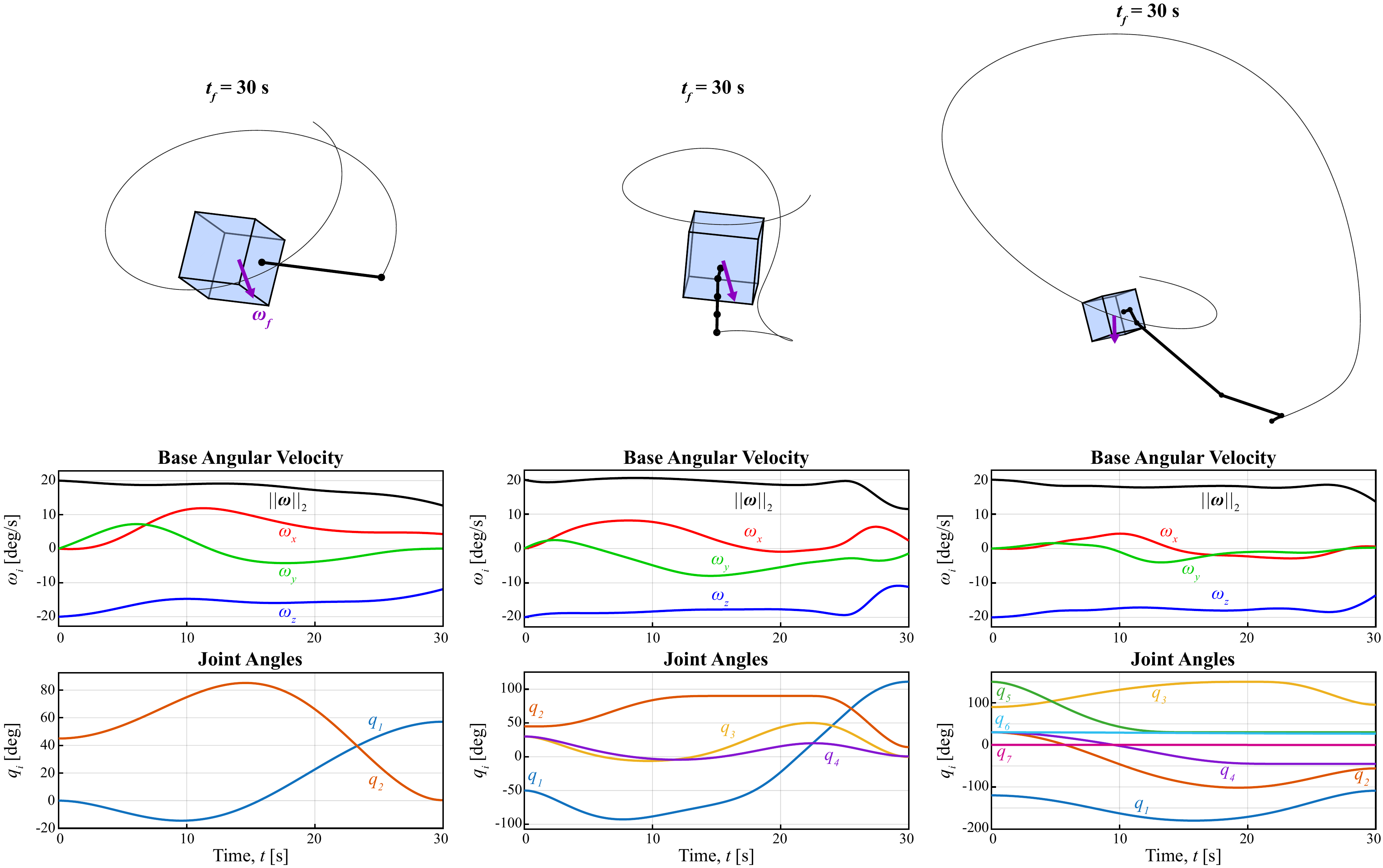}
    \caption{
        Propellant-free detumble maneuvers from $\bm{\omega}_0 = [0,0,-20]~\mathrm{deg/s}$ for the spacecraft-manipulator systems of Fig.~\ref{fig:ManipulatorConfigs}.}
    \label{fig:MegaDetumbleFig}
\end{figure*}

Using this methodology, we compute attitude maneuvers for the free-floating spacecraft-manipulator systems depicted in Fig.~\ref{fig:ManipulatorConfigs}. Here the base spacecraft is modeled as a uniform cube of mass $m_0$ and sidelength $s$, and the manipulators consist of rigid links of mass $m_i$ defined by the DH parameters of Table~\ref{tab:systemparams}. The three systems correspond to (i) a small spacecraft with a 2-DOF, 
single-link manipulator, (ii) a small spacecraft with a 4-DOF manipulator representative of the xLink robotic arm system~\cite{xLink}, and (iii) an ISS-scale spacecraft with a 7-DOF manipulator representative of the  Canadarm2~\cite{nokleby2007singularity}. For each system, we compute an example slew maneuver of $\theta = 20^\circ$ about the axis $\hat{\bm{n}}\approx[0.74,0.02,0.68]$ from a zero initial orientation, and an example detumble maneuver from the initial angular velocity $\bm{\omega}_0 = [0,0,-20]~\mathrm{deg/s}$. A constant initial guess with the state held at $\bm{x}_0$ and controls set to zero is used to initialize each solve. Fig.~\ref{fig:MegaSlewFig} and Fig.~\ref{fig:MegaDetumbleFig} plot the resulting base attitude and joint trajectories computed with a collision clearance of $d_{\mathrm{min}}=0.01~\mathrm{m}$ and maneuver time of $t_f = 30~\mathrm{s}$ (i.e., $N = 600, \Delta t=0.05$). Table~\ref{tab:SolveTimevsDOF} lists the corresponding solve times\footnote{Computed on an Intel Core i9-13900H CPU with 64 GB RAM.}.

The slew maneuvers of Fig.~\ref{fig:MegaSlewFig} demonstrate successful attitude reconfiguration with manipulator motions that avoid collisions and satisfy joint constraints. Due to the smaller mass fraction of the 7-DOF manipulator ($\eta_m=\sum m_i/m_0\approx 3.75e-3$) compared to the 2-DOF and 4-DOF manipulators ($\eta_m\approx8.3e-2$), the 7-DOF manipulator induces relatively smaller torques on the base spacecraft and requires qualitatively larger motions to achieve the same slew maneuver. At the same time, the increased degrees of freedom enable a smoother base attitude trajectory without any overshoot.

The detumble maneuvers of Fig.~\ref{fig:MegaDetumbleFig} demonstrate up to a 40\% decrease in angular velocity solely from manipulator motions. For each spacecraft-manipulator system, the terminal state exhibits a fully extended manipulator to maximize the inertia about the spin axis. This configuration helps decrease the kinetic energy-like terminal cost of Eq.~\ref{eq:cost}, essentially aligning the spin axis with the maximum principal inertia axis of the system.

\begin{figure}[t]
    \centering
    \includegraphics[width=0.675\linewidth]{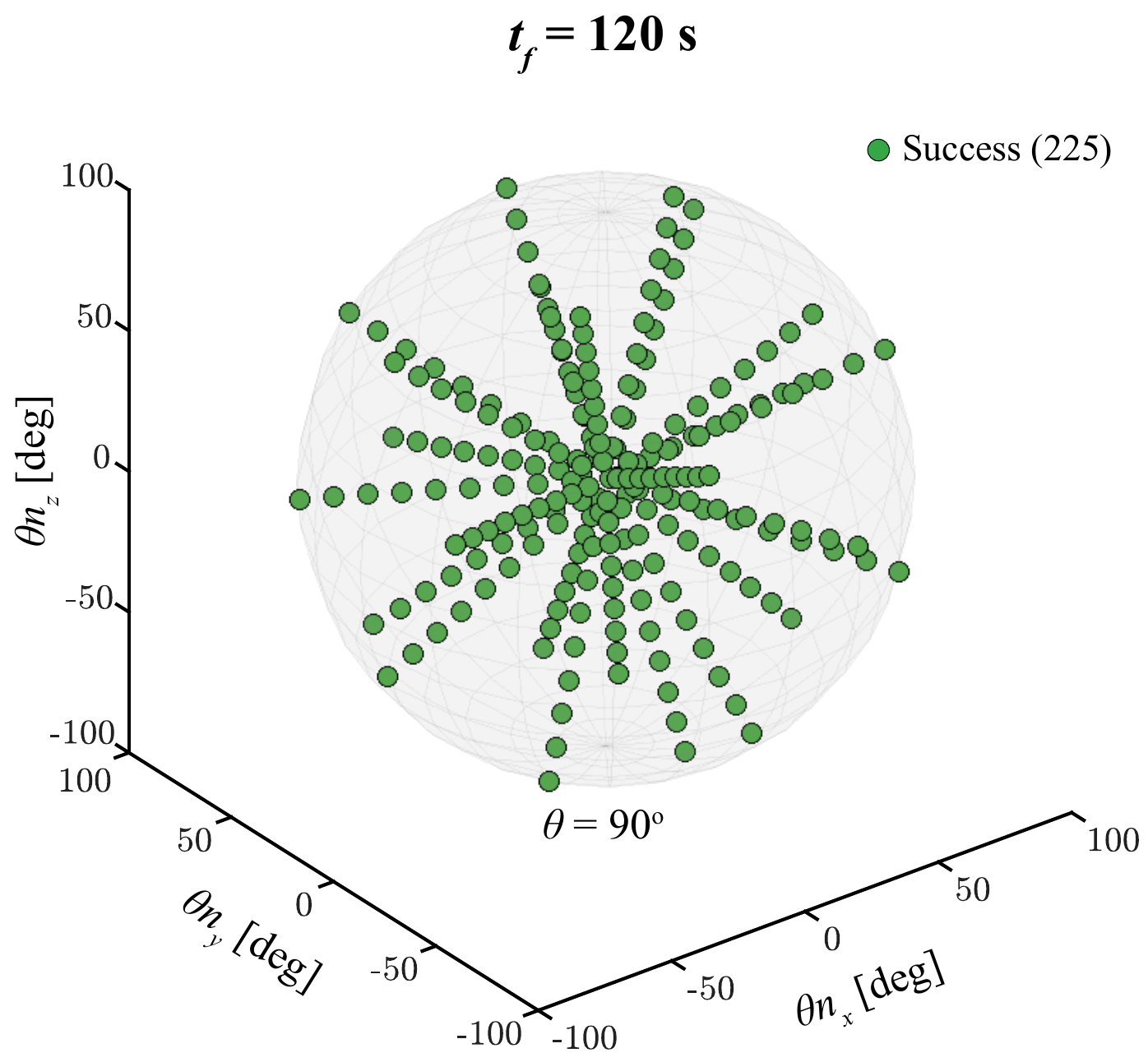}
    \caption{%
        Successful slew maneuvers of $\theta\in[10^\circ,90^\circ]$ with the 2-DOF manipulator about a range of slew axes $\hat{\bm{n}}=[n_x,n_y,n_z]$.} 
    \label{fig:SlewSweepTail}
\end{figure}

To further demonstrate the breadth of attitude maneuvers feasible with the OCP, we compute manipulator motions for a range of slew maneuvers parametrized by the axis-angle representation $\bm{\theta}=\theta\hat{\bm{n}}$. For each slew, we classify whether the maneuver is successful by verifying that the norm of the terminal attitude error satisfies $||\bm{e}_{vf}||_2 \leq 1e-3$. Fig.~\ref{fig:SlewSweepTail} plots the results for the 2-DOF manipulator over a range of slew angles $\theta \in [10^\circ,90^\circ]$, slew axes $\hat{\bm{n}}$ sampled via a spherical Fibonacci mapping~\cite{keinert2015spherical}, and a maneuver time of $t_f = 120~\mathrm{s}$. All 225 attempted maneuvers were classified as successful, demonstrating that manipulator motions achieve slew maneuvers of up to $90^\circ$ about a large range of slew axes within the unit sphere. Experimentally, longer maneuver times and larger manipulator mass fractions correspond to larger achievable slew angles; however, a formal reachability  analysis~\cite{bansal2017hamilton} beyond the scope of this work is required to determine the attitude reachability within $SO(3)$ for arbitrary spacecraft-manipulator system configurations.

Together, the results of Figs.~\ref{fig:MegaSlewFig}-\ref{fig:SlewSweepTail} demonstrate the versatility of the OCP in computing manipulator motions for both slew and detumble trajectories for various real-world space manipulator configurations. While the increasing solve times of Table~\ref{tab:SolveTimevsDOF} highlight the challenges of high-dimensional, nonlinear, constrained search spaces, our methodology achieves successful trajectories even with a simple initialization. Promising future directions to improve solve speed and quality are discussed in Section \ref{sec:conclusion}.

\begin{table}[htp]
\centering
\caption{Solve times for the maneuvers of Figs.~\ref{fig:MegaSlewFig}-\ref{fig:MegaDetumbleFig}.}
\vspace{7pt}
\label{tab:SolveTimevsDOF}
\begin{tabular}{lccc}
\hline
& 2-DOF & 4-DOF & 7-DOF \\
\hline
Slew & 249 s & 361 s & 1318 s \\
Detumble & 279 s & 835 s & 2059 s \\
\hline
\end{tabular}
\end{table}

\section{Comparison With\\ Momentum Exchange Devices}\label{sec:RWcomparison}

The trajectories of Sec.~\ref{sec:attitudemaneuvers} demonstrate attitude control through manipulator motions, such that the manipulator effectively behaves as a momentum exchange device due to the conservation of angular momentum (Eq.~\ref{eq:H}). In this section, we make this analogy more explicit through quantitative comparisons with conventional momentum exchange devices.

\begin{figure}[t]
    \centering
    \includegraphics[width=0.82\linewidth]{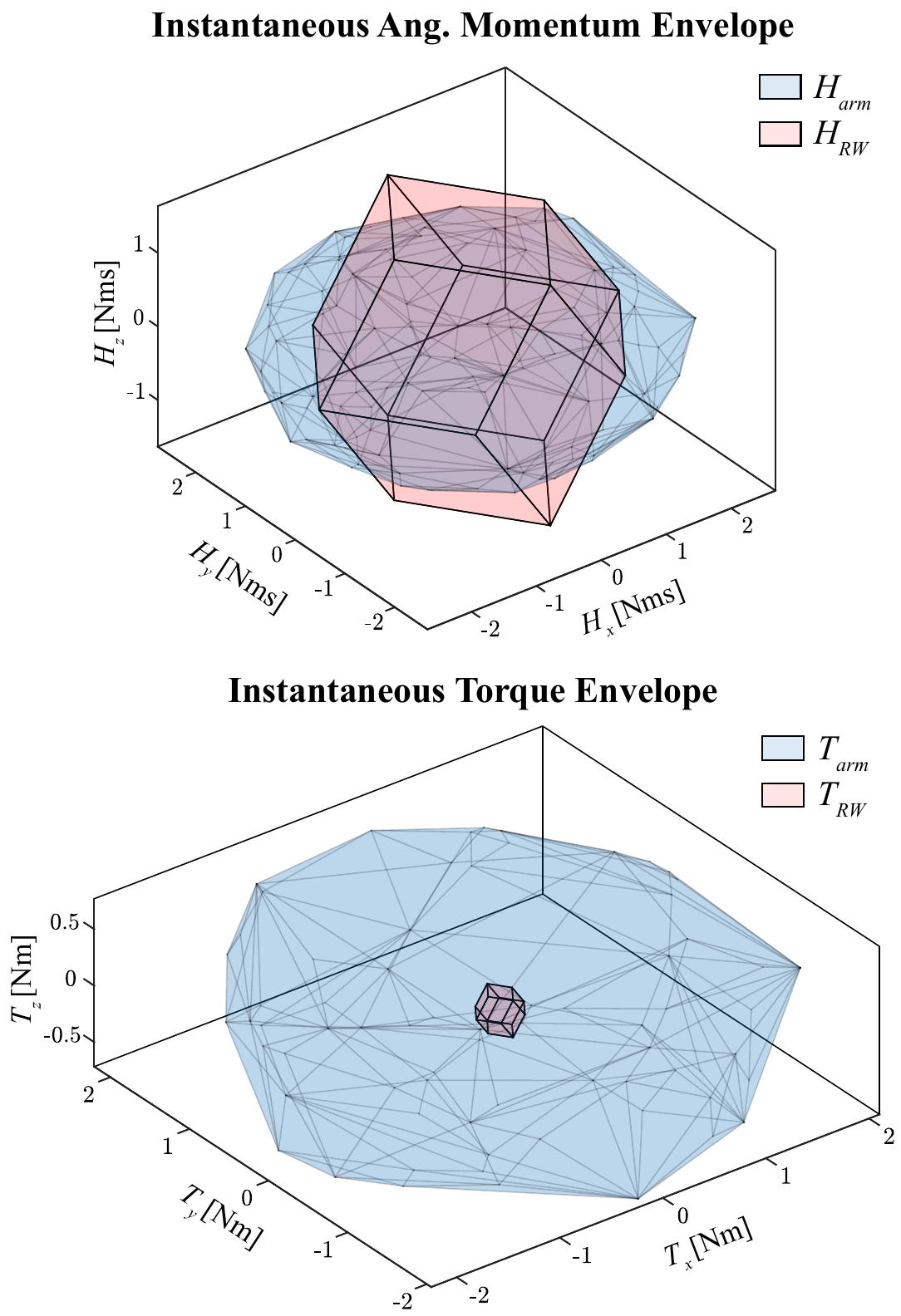}
    \caption{Instantaneous angular momentum and torque envelopes of the 4-DOF manipulator compared with a representative reaction wheel array ($H_{\mathrm{RW}} = 1~\mathrm{Nms}$, $T_{\mathrm{RW}} = 0.1~\mathrm{Nm}$).}
    \label{fig:ManipulatorvsRWEnvelopes}
\end{figure}

While a stationary manipulator cannot store angular momentum in the same manner as spinning reaction wheels or control moment gyroscopes, under the assumption of rigid body dynamics, instantaneous manipulator motion induces angular momentum as described by $\tilde{\bm{H}}(\bm{q},\dot{\bm{q}})$ in Eq.~\ref{eq:H}. Hence, for given joint angles, velocities, and accelerations, we may define the angular momentum and torque associated with manipulator motion as

\vspace{-1em}
\begingroup
\small
\begin{align}
\bm{H}_\mathrm{arm}(\bm{q},\dot{\bm{q}}) &= \sum_{i=1}^{n} \bm{I}_i\bm{\omega}_{i/B} + \sum_{i=1}^{n} m_i [\bm{r}_i-\bm{r}_{CM}]_\times \bm{r}'_i, \label{eq:Harm}\\
\bm{T}_\mathrm{arm}(\bm{q},\dot{\bm{q}},\ddot{\bm{q}}) &= \sum_{i=1}^{n} \left(\bm{I}_i\bm{\omega}_{i/B}\right)' + \sum_{i=1}^{n} m_i [\bm{r}_i-\bm{r}_{CM}]_\times \bm{r}_i'',
\label{eq:Tarm}
\end{align}
\endgroup
where all quantities are expressed in body-frame coordinates and $(\cdot)'$, $(\cdot)''$ denote body-frame time derivatives. Although Eqs.~\ref{eq:Harm}--\ref{eq:Tarm} do not include the gyroscopic terms present in the full equations of motion, they enable a direct comparison with angular momentum and torque envelopes used to characterize momentum exchange devices~\cite{markleyMaximumTorqueMomentum2010}.

For instance, Fig.~\ref{fig:ManipulatorvsRWEnvelopes} plots the convex hull of sampled angular momentum and torque values for the 4-DOF manipulator subject to the joint constraints of Table~\ref{tab:systemparams}. 
Overlaid are the momentum and torque envelopes of a representative four-wheel reaction wheel (RW) array sized for a small spacecraft with mass properties matching those of Table~\ref{tab:systemparams}. Each wheel is assumed to provide a maximum angular momentum storage of $H_{\mathrm{RW}} = 1~\mathrm{Nms}$ and control torque of $T_{\mathrm{RW}} = 0.1~\mathrm{Nm}$, consistent with commercially available small-spacecraft reaction wheels~\cite{RocketLabRW}. The comparison shows that the manipulator provides attitude control authority in all directions similar to the RW array, although the envelope is flatter along the $z$-axis because the manipulator is mounted on that face and generates smaller torques about that axis. Nonetheless, the manipulator achieves a maximum angular momentum and torque of approximately $H_{\mathrm{arm}}\approx 2.7~\mathrm{Nms}$ and $T_{\mathrm{arm}}\approx 2.2~\mathrm{Nm}$, exceeding the representative RW specifications. These results highlight the substantial instantaneous control authority available through manipulator motions, particularly when the manipulator is fully extended. Furthermore, the envelope comparison of Fig.~\ref{fig:ManipulatorvsRWEnvelopes} may be useful for sizing manipulators to achieve attitude control authority comparable to a desired set of momentum exchange devices.

Power consumption presents another useful comparison metric. While the electrical power depends on the specific actuator voltage, current draw, and drive electronics, a lower-bound estimate may be obtained from the mechanical power required for manipulator motion,
\begin{equation}
\label{eq:Parm}
P_{\mathrm{arm}} \approx \frac{1}{\eta}\sum_{i=1}^{j}\tau_i\dot{q}_i,
\end{equation}
where $\tau_i$ and $\dot{q}_i$ are the joint torques and velocities, and $\eta$ represents an overall actuator efficiency. 

\begin{figure}[h!]
    \centering
    \includegraphics[width=0.675\linewidth]{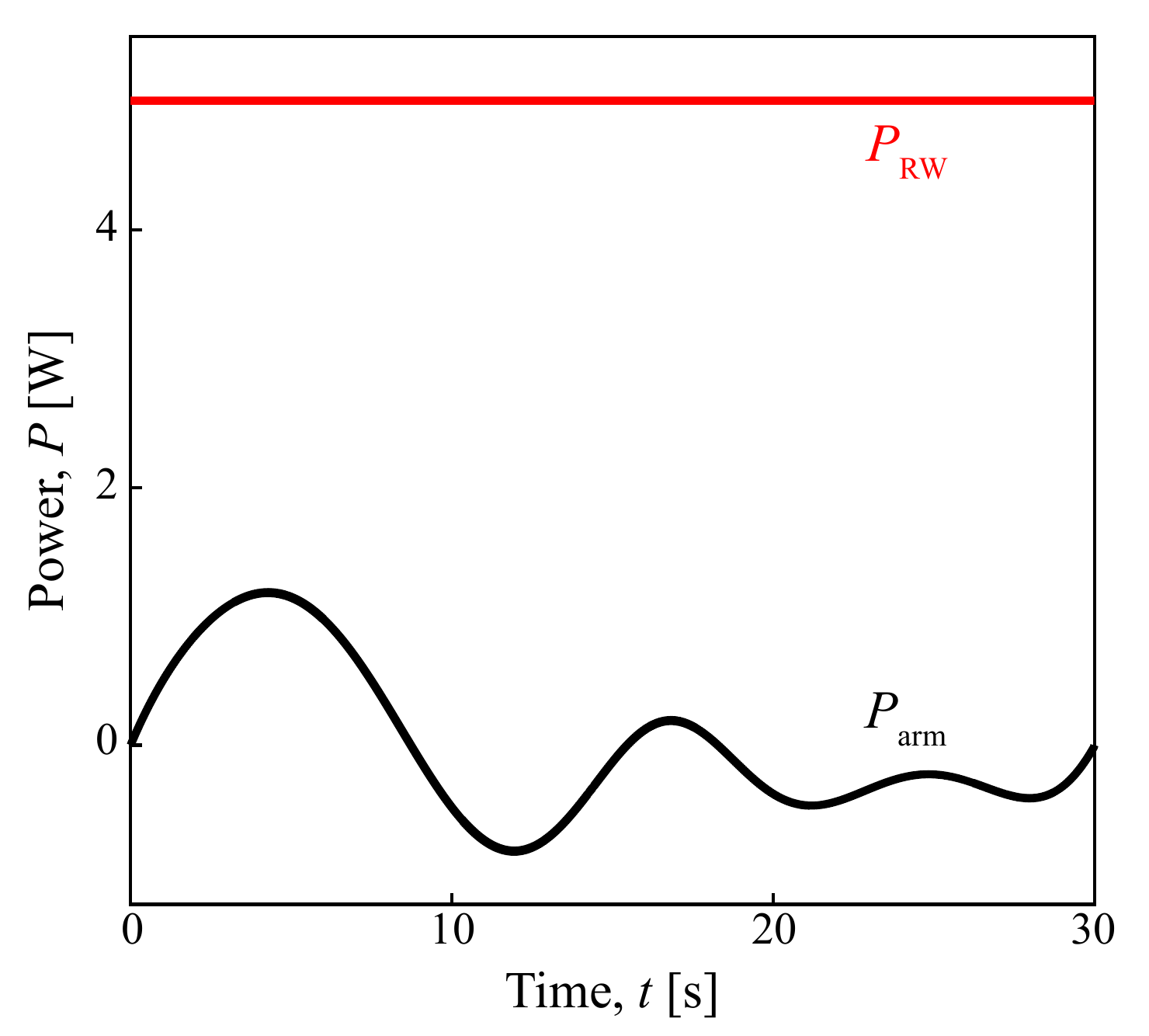}
    \caption{%
        Power consumption for a slew maneuver with the 4-DOF manipulator compared with the steady-state power consumption of a representative reaction wheel ($P_\mathrm{RW} = 5~W$).}
    \label{fig:ManipulatorvsRWPower}
\end{figure}

Fig.~\ref{fig:ManipulatorvsRWPower} plots Eq.~\ref{eq:Parm} for a $20^\circ$ slew maneuver with the 4-DOF manipulator using an assumed efficiency of $\eta=0.1$, together with the steady-state power consumption of a representative reaction wheel, $P_{\mathrm{RW}}=5~\mathrm{W}$~\cite{RocketLabRW}. This value corresponds to nominal operation rather than an active attitude maneuver, and thus represents a lower bound on the power required to perform the same slew using reaction wheels. The manipulator trajectory achieves a peak power consumption of approximately $P_{\mathrm{arm}}\approx1.2~\mathrm{W}$. Although Eq.~\ref{eq:Parm} neglects additional electrical losses, this comparison suggests that manipulator-based attitude maneuvers may require less power than conventional momentum exchange devices. The reduction is attributable to the relatively low angular velocities of the manipulator links compared to the high spin speeds of reaction wheels required for attitude maneuvers.

\section{Conclusion}\label{sec:conclusion}
This paper has presented a numerical trajectory optimization framework for propellant-free attitude control of free-floating spacecraft via manipulator motions. An optimal control problem with joint constraints and collision avoidance was formulated and solved with an interior point method to demonstrate slew and detumble maneuvers for a range of spacecraft-manipulator systems. A comparison with conventional momentum exchange devices also highlighted the potential for larger control torques and lower power consumption with manipulator-based attitude control.

While numerical trajectory optimization has been demonstrated as a viable tool for spacecraft attitude planning using manipulator motions, challenges remain. In particular, the inclusion of self-collision constraints further segments an already non-convex search space, making long-horizon slew and detumble maneuvers increasingly sensitive to initializations and local minima. Future work will investigate informed warm-starting strategies, including geometric attitude interpolations and joint-aware trajectory initializations, to improve both solution quality and computational efficiency for potential real-time use. Learned policies that exploit offline maneuver discovery may also provide a promising approach for planning attitude maneuvers online for constrained, high-DOF spacecraft-manipulator systems.

Additionally, future work will relax the assumption of rigid body dynamics and investigate the effect of flexible vibrations of the manipulator on the desired attitude maneuvers. Further space-environment constraints must also be incorporated, e.g., manipulator motions altering the spacecraft's thermal environment and exposing actuators to different lighting and radiation conditions. Finally, investigating manipulator-based attitude control with larger end effector masses and characterizing the attitude reachability envelope presents an interesting opportunity to expand the range of maneuvers, particularly for in-space assembly and manufacturing applications involving large payload mass fractions.

\addtolength{\textheight}{-5cm}   %

\bibliographystyle{ieeetr}
\bibliography{references}

@article{crain2019experimental,
  title={{Experimental Validation of Pseudospectral-Based Optimal Trajectory Planning for Free-Floating Robots}},
  author={Crain, Alexander and Ulrich, Steve},
  journal={Journal of Guidance, Control, and Dynamics},
  volume={42},
  number={8},
  pages={1726--1742},
  year={2019},
  publisher={American Institute of Aeronautics and Astronautics},
  doi={https://doi.org/10.2514/1.G003528}
}

@article{dubowskyKinematicsDynamicsControl1993,
title = {{The Kinematics, Dynamics, and Control of Free-Flying and Free-Floating Space Robotic Systems}},
  author = {Dubowsky, Steven and Papadopoulos, Evangelos},
  year = 1993,
  journal = {IEEE Transactions on Robotics and Automation},
  issn = {1042296X},
  doi = {10.1109/70.258046},
  isbn = {1042-296X}
}

@inproceedings{dubowsky1991path,
  title={{Path Planning for Space Manipulators to Minimize Spacecraft Attitude Disturbances}},
  author={Dubowsky, Steven and Torres, Miguel A},
  booktitle={Proceedings of IEEE International Conference on Robotics and Automation},
  volume={3},
  pages={2522--2528},
  year={1991},
  organization={Citeseer},
  doi={https://doi.org/10.1109/ROBOT.1991.132005}
}

@book{leveSpacecraftMomentumControl2015,
title = {{Spacecraft Momentum Control Systems}},
  author = {Leve, Frederick A. and Hamilton, Brian J. and Peck, Mason A.},
  year = 2015,
  publisher = {Springer International Publishing},
  address = {Cham},
  doi = {10.1007/978-3-319-22563-0},
  urldate = {2026-04-03},
  copyright = {https://www.springernature.com/gp/researchers/text-and-data-mining},
  isbn = {978-3-319-22562-3 978-3-319-22563-0},
  langid = {english}
}

@article{markleyMaximumTorqueMomentum2010,
title = {{Maximum Torque and Momentum Envelopes for Reaction Wheel Arrays}},
  author = {Markley, F. Landis and Reynolds, Reid G. and Liu, Frank X. and Lebsock, Kenneth L.},
  year = 2010,
  month = sep,
  journal = {Journal of Guidance, Control, and Dynamics},
  volume = {33},
  number = {5},
  pages = {1606--1614},
  issn = {0731-5090, 1533-3884},
  doi = {10.2514/1.47235},
  urldate = {2026-04-03},
  langid = {english}
}

@article{papadopoulosDynamicSingularitiesFree1993,
title = {{Dynamic Singularities in Free- Floating Space Manipulators}},
  author = {Papadopoulos, E and Dubowsky, S},
  year = 1993,
  journal = {Journal of Dynamic Systems, Measurement and Control},
  volume = {115},
  number = {1},
  pages = {44--52},
  isbn = {0022-0434}
}

@article{passerelloHumanAttitudeControl1971,
title = {{Human Attitude Control}},
  author = {Passerello, C.E. and Huston, R.L.},
  year = 1971,
  month = mar,
  journal = {Journal of Biomechanics},
  volume = {4},
  number = {2},
  pages = {95--102},
  issn = {00219290},
  doi = {10.1016/0021-9290(71)90019-4},
}

@article{tortopidisPointtopointMotionPlanning2007,
title = {{On Point-to-Point Motion Planning for Underactuated Space Manipulator Systems}},
  author = {Tortopidis, Ioannis and Papadopoulos, Evangelos},
  year = 2007,
  month = feb,
  journal = {Robotics and Autonomous Systems},
  volume = {55},
  number = {2},
  pages = {122--131},
  issn = {09218890},
  doi = {10.1016/j.robot.2006.07.003},
  urldate = {2026-04-03},
  copyright = {https://www.elsevier.com/tdm/userlicense/1.0/},
  langid = {english}
}

@article{vafaDynamicsManipulatorsSpace1987,
title = {{On the Dynamics of Manipulators in Space Using the Virtual Manipulator Approach}},
  author = {Vafa, Z. and Dubowsky, S.},
  year = 1987,
  journal = {Proceedings. 1987 IEEE International Conference on Robotics and Automation},
  volume = {4},
  pages = {579--585},
  issn = {0278-3649},
  doi = {10.1109/ROBOT.1987.1088032},
  isbn = {0818607874}
}

@article{kane1969method,
title = {{A Method of Active Attitude Control Based on Energy Considerations}},
  author={Kane, TR and Scher, MP},
  journal={Journal of Spacecraft and Rockets},
  volume={6},
  number={5},
  pages={633--636},
  year={1969},
  doi={https://doi.org/10.2514/3.29630}
}

@article{edwards1974automatic,
title = {{Automatic Spacecraft Detumbling by Internal Mass Motion}},
  author={Edwards, Terry L and Kaplan, Marshall H},
  journal={AIAA Journal},
  volume={12},
  number={4},
  pages={496--502},
  year={1974},
  doi={https://doi.org/10.2514/3.49275}
}

@article{watanabe2023attitude,
title = {{Attitude Control and On-Orbit Performance Evaluation of Spacecraft with Variable Shape Function}},
  author={Watanabe, Kei and Kobayashi, Hiroyuki and Amaki, Yuki and Chujo, Toshihiro and Matunaga, Saburo},
  journal={Advances in Space Research},
  volume={72},
  number={6},
  pages={2313--2323},
  year={2023},
  publisher={Elsevier},
  doi={https://doi.org/10.1016/j.asr.2023.06.002}
}

@article{torres1992minimizing,
title = {{Minimizing Spacecraft Attitude Disturbances in Space Manipulator Systems}},
  author={Torres, Miguel A and Dubowsky, Steven},
  journal={Journal of Guidance, Control, and Dynamics},
  volume={15},
  number={4},
  pages={1010--1017},
  year={1992},
  publisher={American Institute of Aeronautics and Astronautics},
  doi={https://doi.org/10.2514/3.20936}
}

@book{lynch2017modern,
title = {{Modern Robotics}},
  author={Lynch, Kevin M and Park, Frank C},
  year={2017},
  publisher={Cambridge University Press}
}

@inproceedings{bansal2017hamilton,
title = {{Hamilton-Jacobi Reachability: A Brief Overview and Recent Advances}},
  author={Bansal, Somil and Chen, Mo and Herbert, Sylvia and Tomlin, Claire J},
  booktitle={2017 IEEE 56th Annual Conference on Decision and Control},
  pages={2242--2253},
  year={2017},
  organization={IEEE},
  doi={https://doi.org/10.1109/CDC.2017.8263977}
}

@article{keinert2015spherical,
title = {{Spherical Fibonacci Mapping}},
  author={Keinert, Benjamin and Innmann, Matthias and S{\"a}nger, Michael and Stamminger, Marc},
  journal={ACM Transactions on Graphics (TOG)},
  volume={34},
  number={6},
  pages={1--7},
  year={2015},
  publisher={ACM New York, NY, USA},
  doi={https://doi.org/10.1145/2816795.2818131}
}

@article{nokleby2007singularity,
title = {{Singularity Analysis of the Canadarm2}},
  author={Nokleby, Scott B},
  journal={Mechanism and Machine Theory},
  volume={42},
  number={4},
  pages={442--454},
  year={2007},
  publisher={Elsevier},
  doi={https://doi.org/10.1016/j.mechmachtheory.2006.04.004}
}

@online{NASAFFR,
  author = {Colleen Wouters},
title = {{NASA’S Fly Foundational Robots Demo to Bolster In-Space Infrastructure}},
  year = {2025},
  url = {https://www.nasa.gov/centers-and-facilities/goddard/nasas-fly-foundational-robots-demo-to-bolster-in-space-infrastructure/},
}

@online{xLink,
  author = {{Motiv Space Systems}},
title = {{xLink Space-Rated Modular Robotic Arm System}},
  year = {2026},
  url = {https://motivss.com/products-capabilities/robotics/xlink/},
}

@online{RocketLabRW,
  author = {{Rocket Lab Corporation}},
title = {{Reaction Wheels}},
  year = {2026},
  url = {https://rocketlabcorp.com/space-systems/satellite-components/reaction-wheels/},
}

@article{wachter2006implementation,
title = {{On the Implementation of an Interior-Point Filter Line-Search Algorithm for Large-Scale Nonlinear Programming}},
  author={W{\"a}chter, Andreas and Biegler, Lorenz T},
  journal={Mathematical Programming},
  volume={106},
  number={1},
  pages={25--57},
  year={2006},
  publisher={Springer},
  doi={https://doi.org/10.1007/s10107-004-0559-y}
}

@article{Andersson2019,
  author = {Joel A E Andersson and Joris Gillis and Greg Horn
            and James B Rawlings and Moritz Diehl},
title = {{{CasADi} -- {A} Software Framework for Nonlinear Optimization and Optimal Control}},
  journal = {Mathematical Programming Computation},
  volume = {11},
  number = {1},
  pages = {1--36},
  year = {2019},
  publisher = {Springer},
  doi = {https://doi.org/10.1007/s12532-018-0139-4}
}

@article{atchison2011length,
title = {{Length Scaling in Spacecraft Dynamics}},
  author={Atchison, Justin A and Peck, Mason A},
  journal={Journal of Guidance, Control, and Dynamics},
  volume={34},
  number={1},
  pages={231--246},
  year={2011},
  doi={https://doi.org/10.2514/1.49383}
}

@article{wie2004solar,
title = {{Solar Sail Attitude Control and Dynamics, Part Two}},
  author={Wie, Bong},
  journal={Journal of Guidance, Control, and Dynamics},
  volume={27},
  number={4},
  pages={536--544},
  year={2004},
  doi={https://doi.org/10.2514/1.11133}
}

@article{kane1969dynamical,
title = {{A Dynamical Explanation of the Falling Cat Phenomenon}},
  author={Kane, Thomas R and Scher, MP},
  journal={International Journal of Solids and Structures},
  volume={5},
  number={7},
  pages={663--670},
  year={1969},
  publisher={Pergamon},
  doi={https://doi.org/10.1016/0020-7683(69)90086-9}
}

@inproceedings{chang2011lizard,
title = {{A Lizard-Inspired Active Tail Enables Rapid Maneuvers and Dynamic Stabilization in a Terrestrial Robot}},
  author={Chang-Siu, Evan and Libby, Thomas and Tomizuka, Masayoshi and Full, Robert J},
  booktitle={2011 IEEE/RSJ International Conference on Intelligent Robots and Systems},
  pages={1887--1894},
  year={2011},
  organization={IEEE},
  doi={https://doi.org/10.1109/IROS.2011.6094658}
}

@article{libby2016comparative,
title = {{Comparative Design, Scaling, and Control of Appendages for Inertial Reorientation}},
  author={Libby, Thomas and Johnson, Aaron M and Chang-Siu, Evan and Full, Robert J and Koditschek, Daniel E},
  journal={IEEE Transactions on Robotics},
  volume={32},
  number={6},
  pages={1380--1398},
  year={2016},
  publisher={IEEE},
  doi={https://doi.org/10.1109/TRO.2016.2597316}
}

@inproceedings{lee2023enhanced,
title = {{Enhanced Balance for Legged Robots Using Reaction Wheels}},
  author={Lee, Chi-Yen and Yang, Shuo and Bokser, Benjamin and Manchester, Zachary},
  booktitle={2023 IEEE International Conference on Robotics and Automation (ICRA)},
  pages={9980--9987},
  year={2023},
  organization={IEEE},
  doi={https://doi.org/10.1109/ICRA48891.2023.10160833}
}

@article{chu2023combining,
title = {{Combining Tail and Reaction Wheel for Underactuated Spatial Reorientation in Robot Falling with Quadratic Programming}},
  author={Chu, Xiangyu and Wang, Shengzhi and Ng, Raymond and Fan, Chun Yin and An, Jiajun and Au, KW Samuel},
  journal={IEEE Robotics and Automation Letters},
  volume={8},
  number={11},
  pages={7783--7790},
  year={2023},
  publisher={IEEE},
  doi={https://doi.org/10.1109/LRA.2023.3322079}
}

@article{rudin2022cat,
  title={{Cat-Like Jumping and Landing of Legged Robots in Low Gravity using Deep Reinforcement Learning}},
  author={Rudin, Nikita and Kolvenbach, Hendrik and Tsounis, Vassilios and Hutter, Marco},
  journal={IEEE Transactions on Robotics},
  volume={38},
  number={1},
  pages={317--328},
  year={2022},
  publisher={IEEE},
  doi={https://doi.org/10.1109/TRO.2021.3084374}
}

@misc{olsen2026lowgravityplanetaryexplorationusing,
title = {{Towards Low-Gravity Planetary Exploration Using Reinforcement Learning for Walking, Jumping, and in-Flight Attitude Control}}, 
      author={Jørgen Anker Olsen and Kostas Alexis},
      year={2026},
      eprint={2605.24643},
      archivePrefix={arXiv},
      primaryClass={cs.RO},
      url={https://arxiv.org/abs/2605.24643}, 
}

@misc{sola2017quaternion,
title = {{Quaternion Kinematics for the Error-State {K}Alman Filter}}, 
  author={Sol\'{a}, Joan},
  year={2017},
  doi={https://doi.org/10.48550/arXiv.1711.02508},
  howpublished={arXiv:1711.02508}
}

@inproceedings{nakamura1990nonholonomic,
  title={{Nonholonomic Path Planning of Space Robots via Bi-Directional Approach}},
  author={Nakamura, Yoshihiko and Mukherjee, Ranjan},
  booktitle={Proceedings., IEEE International Conference on Robotics and Automation},
  pages={1764--1769},
  year={1990},
  organization={IEEE},
  doi={https://doi.org/10.1109/ROBOT.1990.126264}
}
\end{document}